# A Network Intrusions Detection System based on a Quantum Bio Inspired Algorithm

## Omar S. Soliman[1], Aliaa Rassem[2]

*1,2 Faculty of Computers and Information, Cairo University*

**Abstract**: *Network intrusion detection systems (NIDSs) have a role of identifying malicious activities by monitoring the behavior of networks. Due to the currently high volume of networks trafic in addition to the increased number of attacks and their dynamic properties, NIDSs have the challenge of improving their classification performance. Bio-Inspired Optimization Algorithms (BIOs) are used to automatically extract the the discrimination rules of normal or abnormal behavior to improve the classification accuracy and the detection ability of NIDS. A quantum vaccined immune clonal algorithm with the estimation of distribution algorithm (QVICA-with EDA) is proposed in this paper to build a new NIDS. The proposed algorithm is used as classification algorithm of the new NIDS where it is trained and tested using the KDD data set. Also, the new NIDS is compared with another detection system based on particle swarm optimization (PSO). Results shows the ability of the proposed algorithm of achieving high intrusions classification accuracy where the highest obtained accuracy is 94.8 %.*

## 1. Introduction

Intrusion Detection Systems (IDSs) are the systems responsible of identifying malicious activities by monitoring the behavior of users or networks. They detect the abnormal behavior or unwanted traffic and take the appropriate response against it [27]. There are two types of intrusion detection approaches which are the misuse and anomaly detection approaches. Misuse detection uses known patterns, signatures, of unauthorized behavior to detect intrusions. It is quite strong to detect known intrusions but it has low degree of accuracy in detecting unknown intrusions since it relies on signatures extracted by human experts. Anomaly detection establishes a baseline of normal usage patterns and if it finds something that widely deviates from the baseline, the deviation is agged as a possible intrusion. Although it is powerful to identify new types of intrusion as deviations from normal usage, a potential drawback is the high false alarm rate, previously unseen system behaviors may be recognized as anomalies, and hence flagged as potential intrusions [15]. IDSs can also be categorized to host-based and network-based depending on the audit data which they will analyze. Host based systems examine data of users like what files were accessed and what applications were executed. Network-based IDS (NIDS) examines data as packets of information exchange through the networks.

The goal of NIDSs is to quickly and accurately recognize and distinguish the normal and abnormal network connections. They seek to have a high intrusions detection rate and a low false alarms, detecting normal connections as intrusions, to ensure high classification accuracy. There are some challenges facing these systems which make it difficult to achieve such a goal and keep security of networks. These challenges are the current high volume of networks traffic in addition to the increased number of attacks (intrusions) and their complex and dynamic properties. Artificial intelligence and machine learning were used to build different NIDSs but they have shown limitations to achieving high detection accuracy and fast processing times when confronted with these challenges. Computational Intelligence techniques, BIOs, known for their ability to adapt and to exhibit fault tolerance, high computational speed and resilience against noisy information, compensate for the limitations of these two approaches [30].

Many BIOs have been applied as classification techniques for NIDSs, like the artificial immune system (AIS), Artifical Neural Networks (ANNs), particle swarm optimization (PSO) and many others. They were able to easily and automatically extract the the discrimination rules of normal or abnormal behavior from the large networks logs [7]. Also the Quantum Inspired Evolutionary Algorithms (QIEAs) have been used to build NIDSs. QIEAs are some sort of hybrid algorithms appeared in 1990s. They hybridize the classical BIOs with quantum computing (QC) paradigm to improve the performance of these algorithms especially in complex large problems with high dimensions. The algorithms in this new field proved have been used in many applications including networks security where they proved their effectiveness over the traditional BIOs. New hybrid algorithms can be introduced to enhance the performance of QIEAs and increase the quality of obtained solutions.

The aim of this paper is to build a NIDS based on a new proposed BIO named Quantum Vaccined Immune Clonal Algorithm with Estimation of Distribution Algorithm (QVICA-with EDA). This algorithm introduces the vaccine operator and the EDA sampling to be added to the classical Quantum Inspired Immune Clonal Algorithm (QICA). The algorithm is used to build a better NIDS with a higher intrusions classification accuracy. It is trained and tested using the audit data of the benchmark KDD dataset and is compared with another NIDS based on Particle Swarm Optimization (PSO). Results show the superiority of the proposed algorithm over the PSO. The rest of this paper is organized as follows: Section 2 introduces related works in the field of IDS classification algorithms and a background of necessary terms. Section 3 introduces the proposed algorithm where experimental results and discussion are in section 4. The last section is devoted to conclusions and further works.

## 2. Related Works and Background

EAs and QIEAs have been used in past few years in many researches for intrusions classification to increase the detection accuracy of IDS. A sample of these researches is shown in this section. The immune clonal algorithm was used in many networking applications like intrusion detection for securing networks and the spam detection. A Network Intrusion Detection System (NIDS) based on the immunological approach was proposed where an adaptive sampling algorithm was applied during the data collection stage. This algorithm was used according to the dynamic characters of detection data where it was able to improve the data-processing capabilities and the fault tolerance of the system [8]. A collaborative intrusion detection system was proposed to detect denial of service attacks by using the artificial immune system due to its distributional, collaborative, robust and adaptive capabilities. This system was applied for peer to peer networks where it was able to increase the precision of attack discovery and decreases false positive rate [21]. Also, An IDS based on immune algorithm (IA) and support vector machine (SVM) was introduced . In this method, immune algorithm is used to preprocess the network data, SVM is adopted to classify the optimization data, and recognize intruders. Results showed that the feasibility and efficiency of the system [5].

A hybrid intrusion detection system based on rough set (RS )for feature selection and simplified swarm optimization for intrusion data classification was applied. RS is proposed to select the most relevant features and the PSO with a new weighted local search( WLS) strategy was used for classfication. The WLS is added to discover the better solution from the neighborhood of the current solution produced by PSO. The testing results showed that the proposed hybrid system can achieve higher classification accuracy [6]. Particle Swarm Optimization and its variants were combined with various Machine Learning techniques. They were used for Anomaly Detection in Network Intrusion Detection System to enhance the performance of system [22]. An improved negative selection algorithm that integrates a novel further training strategy to reduce self-samples, to reduce computational cost in testing stage, was developed. The algorithm was able to get the highest detection rate and the lowest false alarm rate in most cases [9]. The Support Vector Machine (SVM) was introduced as a classifier for an IDS where a wrapper based feature selection approach using Bees algorithm (BA) as a search strategy for subset generation. The result shows that these combined algorithms has yielded better quality IDS [2]. Also, An improved incremental SVM algorithm (ISVM) combined with a kernel function U-RBF was proposed and applied into network intrusion detection. The simulation results showed that the improved kernel function U-RBF has played some role in saving training time and test time [33]. An anomaly based network IDS using GA approach was adopted where the proposed IDS used an adaptive GA for both learning and detection. The proposed method was efficient with respect to good detection rate with low false positives in addition to the lower execution time [24]. An IDS based on GA was proposed where GA uses evolution theory to information evolution in order to filter the trafic data and so reduce the complexity. It was implemented on KDD99 benchmark dataset and obtained reasonable detection rate [12].

A QIEA using eigenvectors and niching strategy was used to optimize the database of the signature based detection system, an ID system type that is known with its poor detection performance [29], and the algorithm was able to improve the ID's ability in detecting the unknown attacks [34]. A QIEA was also used for optimizing the features selection and kernel parameters of the support vector machine used for anomaly detection [31]. The QGA was also used to optimize the clustering methodology and get the optimal number of clusters to be used for classifying the data collected by the IDS [32]. A quantum neural network (NN) was applied for intrusion application to overcome the weakness of the back propagation NN which may fall into local minimum [4]. The quantum particle swarm optimization (QPSO) was used as a trainer for NN and SVM to get better IDS performance. It was applied to train the wavelet NN to improve the detection rate for anomalies and reduce the false detection alarms in the network anomaly detection [17]. It was also applied with the Gradient Descent (GD) method to train the Radial Basis Function NN for network anomaly detection [18]. It was used with SVM for network Intrusion feature selection and detection where each particle was a selected subset of features and its fitness was defined as the correct classification percentage by SVM [11]. It was used also for solving the linear system of equations of the least square SVM (LS-SVM )to overcome the LS-SVM weakness and guarantee the sparsity and robustness of the solutions [29].

### 2.1. Quantum Inspired Immune Clonal Algorithm

The artificial immune system algorithms (AIS) are a set of EAs inspiring their procedure from the human immune system functionality. AIS include many different algorithms where the two major used algorithms are the negative selection algorithm and the immune clonal algorithm (ICA). ICA is inspired from the human immune systems clonal selection process over the B cells where the evolution process of the antibodies is a repeated cycle of matching, cloning, mutating and replacing. The best B cells are allowed through this process to survive which increases the attacking performance against the unknown antigens. Vaccination is another immunological concept that ICA applies through the vaccine operator. This operator is used to introduce some degree of diversity between solutions and increase their fitness values by using artificial Vaccines.

Quantum inspired evolutionary algorithms (QIEAs) were introduced in the 1990s, they integrate the quantum computing concepts with evolutionary process of EAs. They are able to improve the quality of solutions and enhance the algorithms performance as EAs suffer from bad performance in high dimensional problems. EAs, including the ICA, have to do numerous evolutionary operations and fitness evaluations in large problems which limit them from performing effectively. The hybridization between quantum properties and traditional ICA process enhanced its performance in complex problems. QICA combined quantum computing principles, like quantum bits, quantum superposition property and quantum observation process, with immune clonal selection theory. These concepts are described below in details where the quantum bit representation for antibodies and vaccines in QICA has the advantage of representing a linear superposition of states (classical solutions) in search space probabilistically. Quantum representation can guarantee less population size as a few number of antibodies and vaccines can represent a large set of solutions through the space [23]. The quantum observation process plays a great role in projecting the multi state quantum antibodies into one of its basic states to help in the individuals evaluation.

**-Quantum Bit**: By using quantum bit (q-bit) representation, a small population of antibodies can be created where it represents a larger set of antibodies due to the quantum superpostiion property. The quantum antibody population is initialized with n quantum antibodies using m q-bits for each one. Unlike the classical bit, the q-bit does not represent only the value 0 or 1 but a superposition of the two. Its state can be given by:

$$\psi = \alpha|0> + \beta|1> \qquad (1)$$

Where $\alpha$ and $\beta$ are complex numbers and $\alpha^2$ is the probability to have value 0 and $\beta^2$ is the probability of having value 1 and $\alpha^2 + \beta^2 = 1$.

**- Observation Process:** The quantum represented individuals are converted into binary representation by iterating over each q-bit in the individual and change it to a binary bit. The process is described in algorithm 1

| Algorithm 1 Observation Process |
|---|
| 1: for i = 1 to m do |
| 2:   Generate a random number r between 0 and 1 |
| 3:   *if $i \leq \alpha_i^2$ then* |
| 4:     set the binary bit of i as 0 |
| 5:   else |
| 6:     set the binary bit as 1 |
| 7:   end if |
| 8: end for |

### 2.2. Estimation of Distribution Algorithm

The most of evolutionary algorithms (EA) use sampling during their evolution process for generating new solutions. Some of these algorithms use it implicitly, like Genetic Algorithm (GA), as new individuals are sampled through the genetic operators of the cross over and mutation of the parents. Other algorithms apply an explicit sampling procedure through using probabilistic models representing the solutions characteristics. These algorithms are called the iterated density estimation evolutionary algorithms (IDEAs) where an iterated process of probabilistic model estimation takes place to sample new individuals [10], [16]. Estimation of Distribution Algorithms (EDAs), an example of the IDEA, are population based algorithms with a theoretical foundation on probability theory. They can extract the global statistical information about the search space from the search so far and builds a probability model of promising solutions [25]. Unlike GAs, the new individuals in the next population are generated without crossover or mutation operators. They are randomly reproduced by a probability distribution estimated from the selected individuals in the previous generation [16].

EDA has some advantages, over other traditional EAs, where it is able to capture the interrelations and inter dependencies between the problem variables through the estimation of their joint density function. EDA doesn't have the problem of finding the appropriate values of many parameters as it only relies on the probability estimation with no other additional parameters. The general EDA procedure is shown in Algorithm 2.

---

Algorithm 2 Estimation of Distribution Algorithm

1: Initialize the initial population.
2: while termination condition is not satisfied do
3:   Select a certain number of excellent individuals.
4:   Construct probabilistic model by analyzing information of the selected individuals.
5:   Create new population by sampling new individuals from the constructed probabilistic model.
6: end while

---

The EDA relies on the construction and maintenance of a probability model that generates satisfactory solutions for the problem solved. An estimated probabilistic model, to capture the joint probabilities between variables, is constructed from selecting the current best solutions and then it is simulated for producing samples to guide the search process and update the induced model. Estimating the joint probability distribution associated with the data constitutes the bottleneck of EDA. Based on the complexity of the model used, EDAs are classified into different categories, without interdependencies, pair wise dependencies and multiply dependencies algorithms as below [10] ,

**-Without Interdependencies EDAs:** These models are used when there is no dependency assumed between the variables of the problem. The joint probability distribution is factorized to n independent univariate probability distributions p(xi). Univariate Marginal Distribution Algorithm (UMDA), an example for this category, estimates the p(xi) from the relative marginal frequencies of the Xi of the selected data. Other examples of EDAs under this category are the Population Based Incremental Learning (PBIL) and compact genetic algorithm (cGA) [14]. UMDA, as an example, estimates the $p(x_i)$ from the relative marginal frequencies of the $X_i$ of the selected data.

$$p(x) = \prod_{i=1}^{n} x_i \qquad (2)$$

$$f_N(x, \mu, \in) = \prod_{i=1}^{n} f_N(x_i, \mu_i, \in_i) \qquad (3)$$

**-Pair-wise Dependencies EDAs:** This type of EDAs assumes dependency between pairs of the variables. The joint probability distribution of the variables is factorized as the product of a univariate density function and (n - 1) pair wise conditional density functions given a permutation = $(i_1, \ldots, i_n)$ between variables. Examples are the Bivariate Marginal Distribution Algorithm (BMDA), Mutual Information Maximization for Input Clustering (MIMIC) and Combining Optimizers with Mutual Information Trees (COMIT) algorithms. MIMIC, as an example, searches for the best permutation $\pi^*$, of the n variables, that will minimize the $H_\pi(x)$, Kullback-Leibler divergence function between the $p_\pi(x)$ and p(x) [10].

$$p(x) = p(x_{i1}).p(x_{i2}|x_{i1}) \ldots p(x_{in}|x_{in-1}) \quad (4)$$

**-Multiple Interdependencies EDAs:** Dependencies are assumed between multiple variables where probabilistic graphical models based on either directed or undirected graphs are widely used. Structural and parametric learning are done to learn the topology of the networks and estimate the conditional probabilities. Bayesian network algorithm (BOA), the Markov network EDA and factorized Distribution Algorithm (FDA) are some examples. (For more details, see [14] and [28]).

$$p(x) = \prod_{i=1}^{n} p(x_i|pa(x_i), \theta_i) = \theta_{ijk} \qquad (5)$$

$$f(x) = \prod_{i=1}^{n} N(x_i, \mu_i, v_i) \qquad (6)$$

## 3. Proposed Network Intrusions Detection System

This work builds a NIDS using a new classification algorithm to improve the detection performance. It proposes a new BIO named Quantum Vaccined Immune Clonal Algorithm with Estimation of Distribution Algorithm (QVICAwith EDA). It is based on the quantum computing concepts, immune clonal selection principles and the vaccine operator with EDA sampling. The NIDS is then compared with another system base on PSO [6].A general schema of the proposed NIDS is shown in figure 1. Figure 1 shows that the system has three main stages. First stage is about data preprocessing; the second is the training phase of the proposed classification algorithm and the last one is the testing phase where a detailed description is shown below.

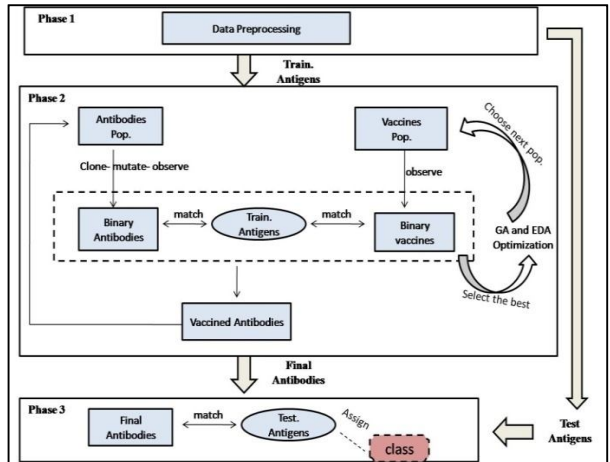

**Fig 1: The new NIDS based on the proposed algorithm (the QVICA-with EDA)**

### 3.1. Data Preprocessing Phase

This phase is concerned with the preprocessing of data represented in the dataset records (network connections). The data are divided into two sets, the training set and the test set. There is a class assigned to each record in the training set, to indicate either it is a normal connection or an attack one, where test records are without class labels. Each record does not have the total 41 features of the KDD, instead it has only six selected features. These six features are those selected and used in the PSO work which are the Service, src_bytes, dst_bytes, Rerror_rate, dst_host_srv_count and dst_host_diff _srv_rate.

The proposed NIDS uses the same selected features for fair comparisons where the symbolic conversion of the symbolic features are done followed by Equal frequency discretization (EFD) for the continuous features. The processed data is the two outputs of this phase where the processed training set, called training Antigens (AGs), is used as an input for the training phase. The processed testing set, called test Antigens, is used as the input for the last phase of testing.

### 3.2. Training Phase

In this phase, the QVICA-V with EDA is trained using the training AGs to learn how to classify the data into normal and intrusions. as mentioned before, the algorithm integrates the quantum computing, for representation, and immune clonal selection principles with the vaccine operator for diversifying solutions. The algorithm also utilizes the EDA probabilistic models and sampling to improve the fitness of antibodies (solutions), increase the degree of diversity and shorten the execution time of the whole algorithm. The main steps of the proposed algorithm are described in Algorithm 3 [23]. As shown in algorithm 3, the algorithm starts by initializing both the quantum antibody population $Q(t)$ and the quantum vaccine population $V(t)$ followed by cloning and mutataing antibodies to be then decoded for evaluation. Additional steps to the simple QICA, like vaccine decoding and sampling will be described in details. The quantum vaccine population $V(t)$ is initialized in the same way with n quantum vaccines where n is the number of grids that the decision space is divided to and $n = (D_1 * D_2 * \dots D_d)$ with d which is the number of dimensions [23].

**- Initialization**: Quantum antibodies and vaccines populations are created where $V(t)$ is initialized with n quantum vaccines where n is the number of grids that the decision space is divided to and $n = (D_1 * D_2 * \dots D_d)$ with d which is the number of dimensions. Quantum antibodies $Q(t)$ are cloned and mutated to get $Q'(t)$ using the clonal operator $\theta$ with C as number of clones that will produce the $Q'(t)$ where $\theta$, C and $Q'(t)$ are as follows,

$$\theta(Q_t) = [\theta(q_1), \theta(q_2), \dots \theta(q_m)] \quad (7)$$

$$C = {N_c}/{n} \quad (8)$$

$$Q'(t) = [Q_t, q'_1, q'_2, \dots, q'_m] \quad (9)$$

Algorithm 3 The proposed Algorithm (QICA-V with EDA)

1: Initialize the quantum antibody and vaccine populations, $Q(t)$ and $V(t)$.
2: Initialize t=1 as first iteration
3: while termination condition is not satisfied do
4:    Apply the clonal and quantum mutation operators over the $Q(t)$ to get $Q'(t)$
5:    Produce B (t) by observing $Q'(t)$.
6:    Decode V (t) to get $V_2$.
7:    Divide $V_2$ into two subpopulations, $V'_2$ and $V''_2$

8:    Select the farthest vaccines from $V'_2$ as the current $V_{best}$.
9:    Estimate probability distribution of the $V_{best}$.
10:   Sample the distribution to get the new $V'_2$
11:   Apply the genetic operators over the $V''_2$ to get the new $V''_2$
12:   Build the new $V_2$ by merging the new $V'_2$ and new $V''_2$
13:   Apply vaccination over B(t) using the new $V_2$ to get BV (t).
14:   Apply clonal selection operator over BV (t) to get Q(t + 1).
15:   end while

**-Vaccine Decoding**: Binary vaccines are converted into decimal representation between 0 and $2^m$ - 1 to get V1 set. These decimal vaccines have to be injected into the decision space so that they are decoded to be within the decision space domain and get V2 set using the following formula,

$$v_i^2 = c_i + \frac{width_j}{2^m} * (v_i^1 - 1)$$
$$and\ i = 1,\dots,n \qquad (10)$$

Where  $c_i$ is the coordinate of the grid (i) and width j is the width of dimension (j) and j = 1,2,…, d. The width can be computed using the maximum and minimum values of each dimension.

**-Vaccine Selection and vaccination**: Hamming distance is used to compute the distance between the vaccines and antibodies to evaluate the farthest vaccines. The Hamming distance function is performed as in algorithm 4.

Vaccines with higher hamming distances from all antibodies are selected (to enhance the exploration) into Vbest set. Vaccines in this set are used to apply the injection process over the mutated AB clones where the injection in our proposed algorithm is done in the real representation.

Algorithm 4 Hamming distnace Function
1: for each bit a in vaccinei and bit b in antibodyj do
2:    if a == b then
3:        mismatchcounter = 0
4:    else
5:        mismatchcounter = 1
6:    end if
7: end for
8: Hamming distance (i,j) = $\sum$ Mismatchcounter

**-Vaccine Sampling**: EDA estimates the probability distribution of the next iterations best vaccines from the current Vbest. It uses the mean and standard deviation (sd), as shown below where b is the length of Vbest, of the vaccines in Vbest to construct its model.

$$\mu = \frac{\sum_{i=1}^b v_i^{best}}{b} \qquad (11)$$

$$\sigma = \sqrt{\frac{\sum[v_i^{best} - \mu]^2}{b}} \qquad (12)$$

**-Clonal selection**: The best antibodies from the vaccined antibodies population and selected to form Q(t+1) and converted again into quantum representation to proceed to a new iteration.

### 3.3. Testing Phase

In this phase, the QVICA-with EDA is tested to evaluate the training process. It is tested using the test antigens, with no class label, produced by the first phase. The final trained antibodies, from the training phase, are matched with the test antigens. Each test antigen is matched with the whole set of these antibodies to be assigned to a class. The higher the number of matched ABs with the antigen, the more probability that this antigen follows their class. At the end of the phase, the class labels of all the test antigens are detected either as normal or attack.

## 4.    Experiments and Results

The proposed algorithm, QVICA- with EDA, is implemented as the classification algorithm for a NIDS and compared with another classification algorithm based on PSO [6]. The experiments were implemented over the The KDD-Cup 99 (Knowledge Discovery and Data Mining Tools Conference) , a benchmark dataset for the netwrok intrusion detection systems [1]. Each record in the KDD represents a TCP/IP connection that is composed of 41 features that are both qualitative and quantitative in nature . There are 39 types of distinct attacks in KDD, grouped into four classes of attack and one class of non attack (normal connections). The main attack types are Denial of Service (DoS),
Probe,Remote-to-Local (R2L) and User-to-Root (U2R) where detailed description of each and its sub types is below [19], [13] and [3].

**-Denial of Service (DoS) attacks:** where an attacker makes some computing or memory resource too busy or too full to handle legitimate requests, thus denying legitimate users access to a machine. Sub attacks of DoS are, Back,Land,neptune,Pod,Smurf and teardrop.
**-Probe attacks:** where an attacker scans a network to gather information or find known vulnerabilities.

Probe's sub attacks are Satan,ipsweep,Nmap and portsweep.

**-Remote-to-Local (R2L) attacks:** where an attacker sends packets to a machine over a network, then exploits machines vulnerability to illegally gain local access as a user. Sub attacks of R2L that can be found in the sets' records are guess-passwd,ftp-write,Imap, Phf,multihop,warezmaster, warezclient and spy.

**-User-to-Root (U2R) attacks:** where an attacker starts out with access to a normal user account on the system and is able to exploit vulnerability to gain root access to the system. U2R sub attacks are Bufier-overow,loadmodule,perl and rootkit .

**Table 1: ACCURACY FOR DIFFERENT EXPERIMENTS USING QVICA-WITH EDA**

| | QVICA-with EDA | | | |
|---|---|---|---|---|
| | Iterations | | | |
| | 10 | 20 | 30 | 40 |
| Pop Size | | | | |
| 10 | 89.0 | 86.5 | 87.7 | 89.9 |
| 20 | 89.1 | 87.1 | 92.5 | 89.1 |
| 30 | 86.6 | 90.8 | 93.1 | 93.4 |
| 40 | 93.8 | 93.2 | 94.7 | 94.5 |
| 50 | 89.3 | 93.4 | 94.8 | 94.2 |

The same parameters settings of the PSO-WLS algorithm are used for our algorithm for fair comparison. A set of 4000 records is selected from the KDD based on the selected features of the PSO-WLS work to evaluate the performance of the QVICA- with EDA. The 10-fold cross validation method is applied where the data are distributed as 10 for testing and the remaining 90 for training.

**Table 2: ACCURACY FOR DIFFERENT EXPERIMENTS USING PSO-WLS**

| | PSO-WLS | | | |
|---|---|---|---|---|
| | Iterations | | | |
| | 10 | 20 | 30 | 40 |
| Pop Size | | | | |
| 10 | 85.6 | 87.0 | 90.4 | 91.8 |
| 20 | 87.7 | 89.0 | 90.6 | 93.2 |
| 30 | 88.6 | 89.5 | 93.5 | 93.5 |
| 40 | 88.6 | 90.0 | 93.5 | 93.6 |
| 50 | 88.6 | 91.5 | 93.5 | 93.6 |

Classification accuracy is the evaluation measure used in this work. The antibodies population size, are set to different values of 10,20,30,40 and 50 where the number of iterations is set to 10,20,30 and 40 with total of 20 independent runs. The best values of accuracy through different experiments are captured and compared with the results of the PSO-WLS algorithm as in tables 1 and 2 where accuracy is as in equation 13 [6],

$$accuracy = \frac{TP + TN}{TP + FP + FN + TN} \quad (13)$$

The tables show that our algorithm outperforms the PSO-WLS algorithm in some experiments where higher accuracy values are obtained. The highest accuracy value was achieved with 50 antibodies and 30 iterations. A sample of the results are also visualized in figures 2, 3 and 4 for more clarification. Using 40 and 50 as the values of the population size, the proposed algorithm is better than the PSO-WLS classification in all different experiments.

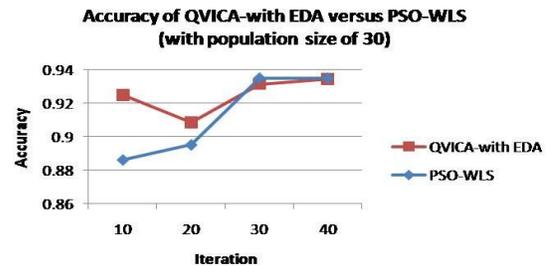

Fig 2:Classification Accuracy at differnt iterations with population size of 30

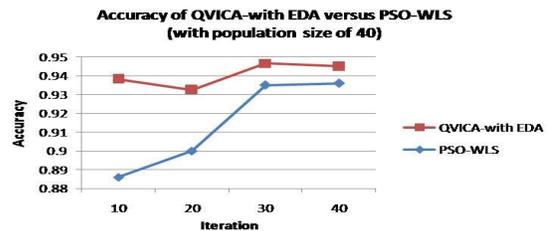

Fig 3: Classification Accuracy at differnt iterations with population size of 40

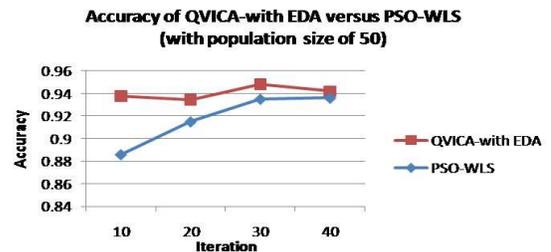

Fig 4: Classification Accuracy at differnt iterations with population size of 50

For more investigation, the accuracy of each fold in one of the best experiments is recorded in table 3 and compared with the folds results of the PSO-WLS algorithm where higher values are obtained by our algorithm. The detailed accuracy values of the 20 runs of the best experiment is in table 4 and are compared with the best experiment of the other algorithm. The proposed algorithm is able to get higher classification accuracy than the best value of the PSO based algorithm using the same parameters.

**Table 3: ACCURACY OF THE 10 FOLD CROSS VALIDATION OF ONE OF THE BEST RUNS**

| Fold | Accuracy Value | |
|------|----------------|---------|
|      | QVICA-with EDA | PSO-WLS |
| 1    | 0.9425 | 0.943 |
| 2    | 0.9776 | 0.925 |
| 3    | 0.9626 | 0.925 |
| 4    | 0.8803 | 0.938 |
| 5    | 0.9476 | 0.933 |
| 6    | 0.9476 | 0.925 |
| 7    | 0.9027 | 0.943 |
| 8    | 0.9651 | 0.945 |
| 9    | 0.9651 | 0.945 |
| 10   | 0.9451 | 0.933 |

**Table 4: ACCURACY OF THE TWO ALGORITHMS OVER 20 RUNS (FOR THE BEST EXPERIMENT)**

| Run | Accuracy Value | |
|-----|----------------|---------|
|     | QVICA-with EDA | PSO-WLS |
| 1   | 94.1 | 93.4 |
| 2   | 92.9 | 93.5 |
| 3   | 93.6 | 93.3 |
| 4   | 94.7 | 93.5 |
| 5   | 94.2 | 93.2 |
| 6   | 92.6 | 93.3 |
| 7   | 88.0 | 93.2 |
| 8   | 90.2 | 93.4 |
| 9   | 91.2 | 93.5 |
| 10  | 94.0 | 93.4 |
| 11  | 92.1 | 92.5 |
| 12  | 92.3 | 93.3 |
| 13  | 93.9 | 93.2 |
| 14  | 91.8 | 93.4 |
| 15  | 93.3 | 93.3 |
| 16  | 93.8 | 93.0 |
| 17  | 92.0 | 93.4 |
| 18  | 91.1 | 93.5 |
| 19  | 94.8 | 93.0 |
| 20  | 89.6 | 93.4 |
| Mean | 94.81 | 93.3 |

## 5. Conclusions

A quantum vaccined immune clonal algorithm with the estimation of distribution algorithm (QVICA-with EDA) was proposed in this paper as a classification algorithm for the NIDS. It was compared with another classification algorithm based on particle swarm optimization (PSO) on the KDD data set. Classification accuracy values obtained at the different experiments showed the ability of the algorithm of achieving high classification accuracy. It outperforms the other algorithm in many experiments which proved its effectiveness. More experiments with different parameters settings will be done in future work to ensure that the algorithm is powerful for intrusion detection.